\documentclass[runningheads]{llncs}
\usepackage{caption}
\usepackage{subcaption}
\usepackage{todonotes}
\usepackage{soul}
\usepackage[T1]{fontenc}
\usepackage{graphicx}
\usepackage{color}
\usepackage[hyphens]{url}
\usepackage{hyperref}
\usepackage{amsmath}
\usepackage{todonotes}
\begin{document}
\title{Scene Text Detection and Recognition "in light of" Challenging Environmental Conditions using Aria Glasses Egocentric Vision Cameras}
\titlerunning{STDR Aria Glasses}

% anonymous for double blind submission

\author{Joseph De Mathia\inst{1} \and
Carlos Francisco Moreno-García\inst{1}\orcidID{0000-0001-7218-9023}}
%Third Author\inst{3}\orcidID{2222--3333-4444-5555}}
%
\authorrunning{De Mathia et al.}
\institute{Robert Gordon University, Aberdeen, AB10 7GJ, UK
%Springer Heidelberg, Tiergartenstr. 17, 69121 Heidelberg, Germany
\email{c.moreno-garcia@rgu.ac.uk}\\
\url{http://www.cfmgcomputing.blogspot.com}}
\maketitle              % typeset the header of the contribution
\begin{abstract}

In an era where wearable technology is reshaping applications, Scene Text Detection and Recognition (STDR) becomes a straightforward choice through the lens of egocentric vision. Leveraging Meta’s Project Aria smart glasses, this paper investigates how environmental variables, such as lighting, distance, and resolution, affect the performance of state-of-the-art STDR algorithms in real-world scenarios. We introduce a novel, custom-built dataset captured under controlled conditions and evaluate two OCR pipelines: EAST with CRNN, and EAST with PyTesseract. Our findings reveal that resolution and distance significantly influence recognition accuracy, while lighting plays a less predictable role. Notably, image upscaling emerged as a key pre-processing technique, reducing Character Error Rate (CER) from 0.65 to 0.48. We further demonstrate the potential of integrating eye-gaze tracking to optimise processing efficiency by focusing on user attention zones. This work not only benchmarks STDR performance under realistic conditions but also lays the groundwork for adaptive, user-aware AR systems. Our contributions aim to inspire future research in robust, context-sensitive text recognition for assistive and research-oriented applications, such as asset inspection and nutrition analysis. The code is available at \url{https://github.com/josepDe/Project_Aria_STR}.

\end{abstract}

%%%%%%%%%%%%%%%%%%%%%%%%%%%%%%%%%%%%%%%%%%%

\section{Introduction}
\label{intro}

Scene Text Detection and Recognition (STDR) is a long-standing problem within the Document Analysis and Recognition (DAR) community. It deals with the detection and classification of characters found in natural images, including pictures taken from advertisements, signposts, books, among others~\cite{Chen2021}. Recently, Meta launched their Project Aria Research initiative~\cite{aria}, consisting of glasses with multiple cameras and sensors. These glasses have been released to the scientific community to perform innovative research on how to improve Virtual Reality (VR) and Augmented Reality (AR) applications~\cite {Plizzari2024}, but also aid in the creation of datasets and algorithms based on egocentric vision in order to improve robotics~\cite{Grauman2024},~\cite{Liu2025} imitation learning~\cite{Hussein2017}, amongst others.

Application-wise, Project Aria’s glasses could help bolster applications in numerous real-life fields where there is a need to inspect an asset~\cite{Toral2023} or to understand an individual’s emotion through facial analysis~\cite{Wired-emotions2024}. In our work, we explore how these glasses can assist nutrition experts in understanding consumers dietary habits. Specifically by analysing the video feed recorded by a participant and understanding their eye gaze behaviour when buying a product; which information they are focusing on, their dietary patterns (e.g quantities and portions) and their overall interaction with food products.

The aim of this project is to study STDR specifically using footage captured by Project Aria glasses. The study focuses on how environmental and image quality factors such as lighting, distance, and resolution impact the accuracy of STDR algorithms. For this study, we have used the Efficient and Accurate Scene Text detection(EAST) algorithm~\cite{Zhou2017} to detect the text bounding boxes. Subsequently, we implement a heuristic correction to merge individual character bounding boxes together, thus conforming the word areas. For the Optical Character Recognition (OCR) stage, this study utilises two different algorithms: Google’s Pytesseract~\cite{pytesseract} and a Convolutional Recurrent Neural Network (CRNN) provided through EasyOCR~\cite{EasyOCR}. Following the convention set at the International Conference on Document Analysis and Recognition (ICDAR) 2024 competition~\cite{Jahagidar2024}, in this project we will evaluate our methods based on the Character Error Rate (CER). To understand how lighting, distance and resolution affect the OCR models, we collated a custom dataset by using the glasses. This dataset contains images of a ground truth poster in different lighting conditions and at varying recorded distances and resolutions.

The rest of the paper is organised as follows. Section \ref{related} presents the related work to egocentric vision in the context of STDR. Section \ref{methods} describes our methodological approach. Section \ref{exps} discusses our three experimental validations, with Section \ref{concs} concluding the report and pointing out future research directions.

% These algorithms will first be run without any unnecessary pre-processing to ascertain the impact that the environmental variables have on their performance, the findings of this step and an attempt at pre-processing to mitigate these issues will be performed.
% A second evaluation of the algorithms with the new pre and post-processing techniques will be carried out on their previous datasets to evaluate whether any improvements have occurred. Subsequently the most effective algorithm identified during evaluation will be selected for real-world testing and further qualitative assessment.

%%%%%%%%%%%%%%%%%%%%%%%%%%%%%%%%%%%%%%%%%%%

\section{Related Work}
\label{related}

Jahagirdar et al~\cite{Jahagidar2024} introduced the ICDAR 2024 Competition on Reading Documents Through Aria Glasses, aimed at advancing STR (Scene Text Recognition) in challenging scenarios characterised by low resolution, low light, and egocentric perspectives. The competition utilised the RDAG-1.0 dataset, collected via Meta’s Project Aria glasses, and included three tasks: isolated word recognition, reading order prediction, and page-level recognition. The paper outlines the OCR approaches submitted, emphasising those that achieved top performance, including transformer-based architectures and hybrid CNN-Transformer pipelines. Critically, this competition served as a foundational reference for this project. It provides standardised benchmarks and highlights practical limitations and strengths of state-of-the-art methods in real-world contexts. It guides our evaluation strategy for Scene Text Recognition (STR) using Aria captured data and setting baseline expectations for the performance of the algorithms applied in this paper.

Mucha et al~\cite{Mucha2024} present a system combining Project Aria smart glasses with a Large Language Model (LLM) to support individuals with visual impairments in real-world reading tasks. Utilising Project Aria Glasses RGB cameras, the system captures egocentric video, performs OCR to extract text, and then processes the recognised content using GPT-4. This approach enables natural language interactions with textual content, such as menu items, resulting in a recognition accuracy of 96.77

To our knowledge, this is the first time that environmental conditions have been systematically studied to understand the potential of egocentric vision in STDR. Furthermore, we are not aware of previous work utilising both RGB and gaze tracking cameras to map where in the scene an individual is looking, with the aim of focusing the text detection.

%%%%%%%%%%%%%%%%%%%%%%%%%%%%%%%%%%%%%%%%%%%

\section{Methodology}
\label{methods}

To study Scene Text Detection and Recognition (STDR) in egocentric contexts, this project will investigate how environmental and image quality factors, specifically lighting, distance, and resolution will impact the accuracy of STDR on footage captured using Project Aria Glasses. A custom dataset will be collected utilising the Project Aria Glasses in order to support controlled experimentation. This approach will allow for precise control over lighting conditions, resolution settings, and camera-to-text distance. Both text detection and recognition will be performed on this dataset.

A selection of STDR algorithms  will be made based on key factors such as performance, specificity to the task and implementation feasibility. 

Text recognition accuracy will be evaluated utilising, Character Error Rate (CER) following the evaluation framework established at the ICDAR 2024 Competition on Reading Documents Through Aria Glasses \cite{Jahagidar2024}.The evaluation of text Recognition algorithms will be carried out on the custom dataset to properly assess the effect of environmental variables, distance, and resolution.

For text detection, Intersection over Union (IOU) will be used as the primary evaluation metric. The evaluation of text detection will be carried out on a small subset of the competition dataset, as this contains practical annotation of ground truth bounding boxes which will be unavailable in the custom dataset.

Based on insights gained from the initial evaluations, a set of preprocessing techniques will be explored to mitigate the adverse effects of lighting, distance, and resolution. These techniques will then be re-applied to the original datasets to determine whether they yield performance improvements.

Finally  the most effective algorithm identified during this evaluation process will be selected for testing, using Project Aria Glasses footage recorded in real world scenarios, such as food shopping in a supermarket.

\begin{figure}[t]
\centering
\includegraphics[width=0.55\textwidth]{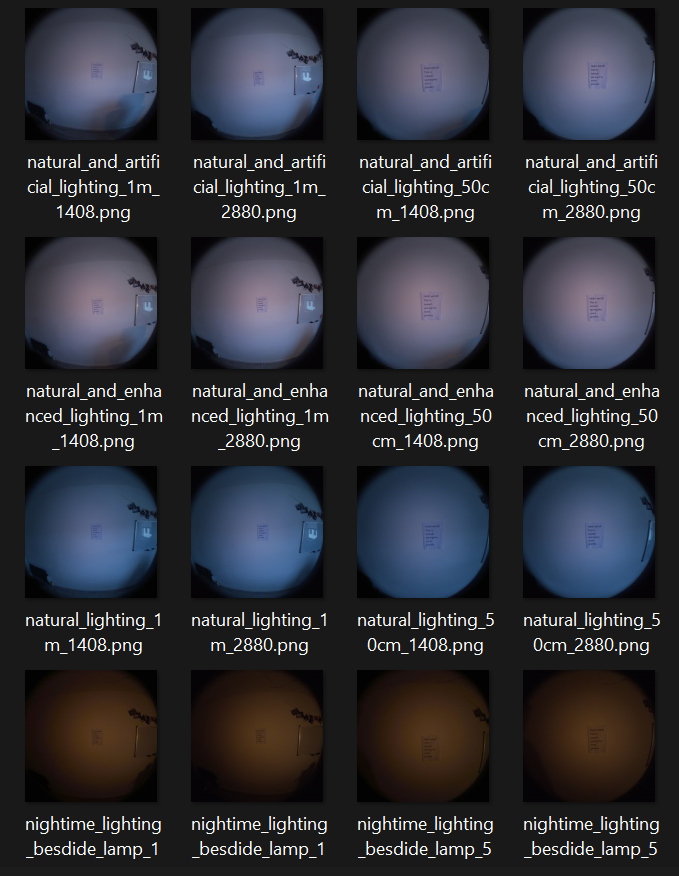}
\caption{Proprietary images captured under different conditions for validation.} 
\label{db}
\end{figure}

\subsection{Proprietary dataset creation}
\label{datasetcreation}
A custom dataset will be created specifically for this project using the Project Aria Glasses.  Multiple videos will be recorded of a poster affixed on a white background wall with the fixed text of \textit{“Hello world! This is Joseph testing the Meta glasses”}. 
The videos will be recorded in four different lighting conditions:
\begin{itemize}
    \item Natural Lighting: recorded at noon on a bright but overcast day, with no artificial light sources.
    \item Natural and Artificial Lighting: recorded under the same natural conditions with the addition of a bedside lamp.
    \item Natural and Enhanced Artificial Lighting: recorded at noon with both an overhead light and a bedside lamp.
    \item Night-Time Lighting: recorded at night, using only a bedside lamp for illumination.
\end{itemize}
For each lighting condition, two separate videos will be captured: one with the glasses positioned approximately 50 cm from the poster, and another at approximately 1 m distance. Additionally, utilising custom recording profiles on the Companion App, each recording will be repeated at two different resolutions: 1048×1408 and 2880×2880, as supported by the glasses.
All samples are shown in Figure \ref{db} \newline

This systematic approach will enable the creation of a controlled dataset that varies key factors such as lighting, distance, and image resolution, allowing for a detailed analysis of their impact on STDR.
The ICDAR 2024 Competition on Reading Documents Through Aria Glasses \cite{Jahagidar2024} dataset will also be utilised for evaluation purposes. Recordings of real-life scenarios (e.g. shopping)  will also be captured with a recording profile of 2880 $\times$ 2880 at 20 frames per second with the use of the internal eye-tracking cameras.

\subsection{Algorithm Selection and Baseline Evaluation}
\label{algselect}

For text detection, this project will employ the Efficient and Accurate Scene Text Detector (EAST) algorithm. Trained on the ICDAR 2015 competition dataset, achieving an F-score of 0.7820~\cite{Zhou2017}, EAST outperformed many existing models despite its lightweight design. It offers high processing speed, achieving up to 13.2 frames per second on 720p images. Although the images captured with the Project Aria Glasses are of higher resolution (1408×1408 and 2880×2880), EAST remains a suitable choice for this project due to its efficiency and ability to detect text at a wide range of orientations. These attributes make EAST particularly well-suited for dynamic, egocentric recordings where lighting conditions, camera angles, and scene compositions vary significantly. While other text detection algorithms, such as PyTesseract, were considered, EAST was ultimately chosen for its robustness, efficiency, and prior success in comparable tasks.

In this project, a heuristic approach will be used to merge the bounding boxes detected by the EAST algorithm. Bounding boxes will be grouped based on their vertical and horizontal proximity. First, bounding boxes are sorted according to their Y-coordinate, grouping text elements likely positioned on the same line. The boxes are then grouped based on two main conditions: vertical proximity, where the top or bottom edges are within a predefined threshold (epsilon\_y), and horizontal separation, where horizontally distant boxes are treated as belonging to separate lines or columns.

After grouping, each set of bounding boxes will be  merged into a single bounding box, calculated using the minimum and maximum X and Y coordinates within the group. This process continues iteratively until all boxes are processed, resulting in a set of merged bounding boxes that best represent coherent text regions. This approach aims to combine spatially related boxes in order to capture complete text units more accurately in lines. 

Following detection and grouping, the identified complete text regions will be passed to text recognition algorithms for further evaluation. Two OCR Models were selected for this purpose:
\begin{itemize}
    \item CRNN - A pre-trained CRNN from the EasyOCR library was used for its ability to maintain performance in suboptimal imaging conditions due to its resilience to distortions and noise.
    \item PyTesseract - A Python wrapper for Google's Tesseract-OCR engine, was chosen as a secondary method due to its ease of integration, extensive documentation, lightweight nature, and support for multilingual text extraction.
\end{itemize}

\begin{figure}[t]
\centering
\includegraphics[width=0.85\textwidth]{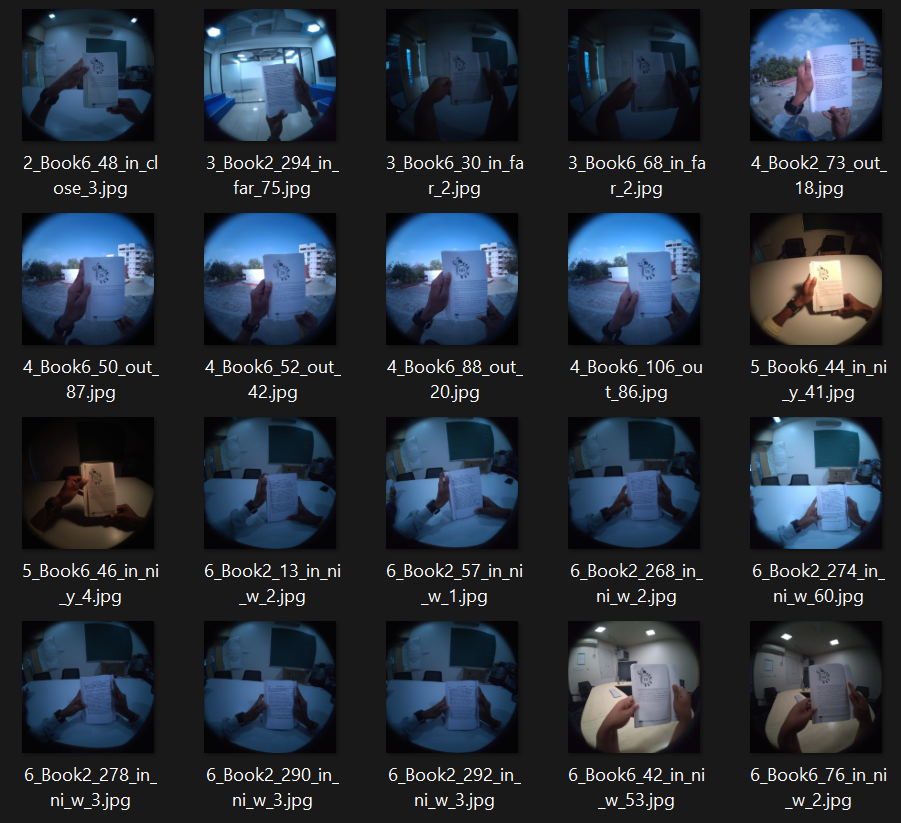}
\caption{Images used from the ICDAR 2024 dataset for validation.} 
\label{db2}
\end{figure}

\subsection{Environmental Impact Analysis and Mitigation}
\label{sec:methodsenviron}
This project will assess the impact of lighting, distance, and resolution on the performance of text recognition and detection algorithms. To achieve this, different lighting metrics will be extracted from the gathered images and their respective impacts will be evaluated. The same process will be executed for distance and resolution.

After testing, pre-processing techniques will be implemented in accordance with the findings of the first evaluation step, with the goal of improving the baseline results and achieving better overall performance. The algorithms will then be re-evaluated with the applied pre-processing techniques on their respective datasets. After evaluating the performance of all the algorithms with the respective processing techniques, the best performing algorithm will be selected for real world testing scenarios.

To evaluate the project's STDR methods in real-world conditions, a recording will be made in a public environment rich in text. This recording will utilise the RGB camera configured to capture egocentric footage at an appropriate resolution and a frame rate of 20 frames per second. In addition, the eye-tracking cameras will be activated to enable Aria’s eye-tracking services. The goal of real-world testing is to evaluate the algorithm’s performance and feasibility in real contexts and to identify its limitations.

%%%%%%%%%%%%%%%%%%%%%%%%%%%%%%%%%%%%%%%%%%%%%

\section{Experimental Validation}
\label{exps}

\subsection{Validation of selected methods using publicly available dataset}
\label{sec:exp1}

A subsection of the ICDAR 2024 Competition on Reading Documents Through ARIA Glasses dataset was used for initial evaluation. Specifically, 20 randomly selected images were extracted from the \textit{Training\_Set1} section of the dataset. This selection was driven by time and computational resource constraints. The ICDAR 2024 dataset contains diverse images captured using Project Aria Glasses and includes corresponding ground truth annotations in the form of JSON files, specifying bounding boxes for textual content. Due to the dataset’s structure, reliable methods for retrieving distance or resolution metadata were unavailable. Therefore, the evaluation focused only on assessing the impact of lighting variations on text detection performance. Results are shown in Table \ref{table}.

\begin{table}[t]
\caption{Results from experimenting with the EAST model for text detection in 20 samples from the ICDAR 2024 datasets which resemble different conditions to be experimented in the proprietary dataset. The highest values for each column are highlighted in \textbf{bold}.}
\label{table}
\begin{tabular}{l|llll|lll}
Image Name                   & \begin{tabular}[c]{@{}l@{}}Mean\\ Bright\\ (lumens)\end{tabular} & \begin{tabular}[c]{@{}l@{}}Std.\\ Bright\\ (lumens)\end{tabular} & \begin{tabular}[c]{@{}l@{}}Global\\ Bright\\ (lumens)\end{tabular} & Contrast     & Prec.         & Rec.          & F1            \\ \hline
6\_Book6\_42\_in\_ni\_w\_53  & 93.97                                                            & 58.34                                                            & 102.92                                                             & \textbf{255} & 0.84          & 0.63          & 0.72          \\
3\_Book2\_294\_in\_far\_75   & 88.07                                                            & 56.68                                                            & 122.26                                                             & \textbf{255} & 0.83          & 0.55          & 0.66          \\
5\_Book6\_44\_in\_ni\_y\_41  & 81.02                                                            & \textbf{68.48}                                                   & 94.93                                                              & \textbf{255} & 0.91          & 0.72          & 0.81          \\
4\_Book6\_52\_out\_42        & 80.48                                                            & 53.09                                                            & 124.09                                                             & \textbf{255} & 0.78          & 0.59          & 0.67          \\
6\_Book6\_76\_in\_ni\_w\_2   & 53.90                                                            & 38.51                                                            & 59.63                                                              & \textbf{255} & 0.92          & 0.70          & 0.80          \\
4\_Book6\_88\_out\_20        & 82.98                                                            & 53.37                                                            & 128.06                                                             & \textbf{255} & 0.82          & 0.69          & 0.75          \\
4\_Book6\_50\_out\_87        & 76.93                                                            & 50.94                                                            & 119.89                                                             & \textbf{255} & 0.79          & 0.61          & 0.69          \\
6\_Book2\_13\_in\_ni\_w\_2   & 44.06                                                            & 29.15                                                            & 68.44                                                              & 240          & 0.61          & 0.42          & 0.50          \\
2\_Book6\_48\_in\_close\_3   & 50.82                                                            & 32.71                                                            & 73.71                                                              & \textbf{255} & \textbf{0.96} & 0.68          & 0.80          \\
6\_Book2\_274\_in\_ni\_w\_60 & \textbf{99.85}                                                   & 58.25                                                            & 138.32                                                             & \textbf{255} & 0.88          & 0.57          & 0.69          \\
3\_Book6\_30\_in\_far\_2     & 25.10                                                            & 20.05                                                            & 37.96                                                              & 194          & 0.82          & 0.33          & 0.47          \\
6\_Book2\_278\_in\_ni\_w\_3  & 42.78                                                            & 30.34                                                            & 66.78                                                              & 252          & 0.87          & 0.70          & 0.77          \\
3\_Book6\_68\_in\_far\_2     & 25.52                                                            & 19.61                                                            & 38.64                                                              & 190          & 0.86          & 0.14          & 0.24          \\
4\_Book6\_106\_out\_86       & 86.19                                                            & 54.77                                                            & \textbf{133.69}                                                    & 245          & 0.83          & 0.65          & 0.73          \\
5\_Book6\_46\_in\_ni\_y\_4   & 27.17                                                            & 30.55                                                            & 34.25                                                              & 183          & 0.86          & 0.67          & 0.75          \\
6\_Book2\_292\_in\_ni\_w\_3  & 43.94                                                            & 29.87                                                            & 68.56                                                              & 251          & 0.95          & \textbf{0.77} & \textbf{0.85} \\
6\_Book2\_268\_in\_ni\_w\_2  & 43.80                                                            & 29.82                                                            & 68.42                                                              & 251          & 0.79          & 0.54          & 0.64          \\
6\_Book2\_290\_in\_ni\_w\_3  & 44.10                                                            & 30.38                                                            & 68.76                                                              & 253          & 0.93          & 0.73          & 0.81          \\
6\_Book2\_57\_in\_ni\_w\_1   & 61.55                                                            & 39.67                                                            & 92.55                                                              & 253          & 0.52          & 0.38          & 0.44          \\
4\_Book2\_73\_out\_18        & 92.63                                                            & 55.80                                                            & 132.72                                                             & 251          & 0.79          & 0.56          & 0.66         
\end{tabular}
\end{table}

\begin{figure}[t]
\includegraphics[width=\textwidth]{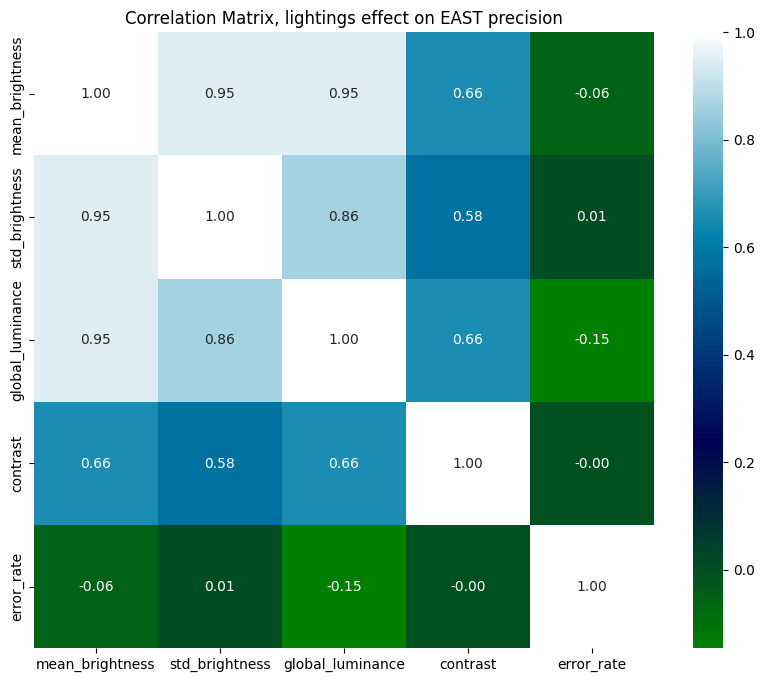}
\caption{Proprietary images captured under different conditions for validation.} 
\label{corrmat}
\end{figure}

\begin{figure}[t]
\centering
\includegraphics[width=0.3\textwidth]{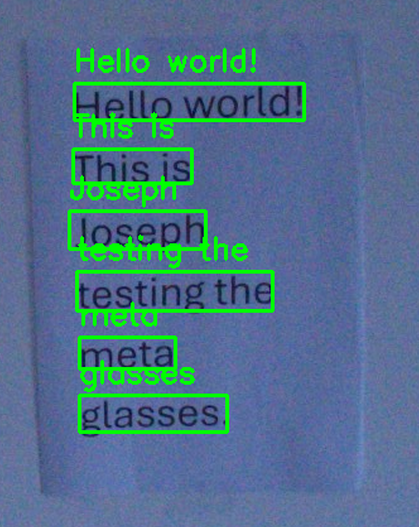}
\caption{An example of text detected and recognised on our dataset images.} 
\label{textdetected}
\end{figure}

Using an Intersection over Union (IOU) threshold of 0.50, the EAST text detection algorithm achieved a mean F1 score of 0.67, a mean precision score of 0.82, and a mean recall of 0.5, on the selected subset (see Figure \ref{db2}). Further analysis was conducted by examining five lighting-related factors: mean brightness, standard deviation of brightness (STD\_brightness), global luminance, image contrast, and error rate. As shown in the correlation matrix (see Figure \ref{corrmat}), lighting variables such as mean brightness, global luminance, and contrast exhibited minimal correlation with recognition error rate (all correlations |r| $\leq$ 0.15 ).

\subsection{Validation of selected methods using proprietary dataset}
\label{sec:exp2}

For text detection and recognition, both the CRNN and PyTesseract algorithms were evaluated on the custom dataset, which enabled assessment across a range of distances, resolutions, and lighting variables. The lighting characteristics analysed included mean brightness, standard deviation of brightness (std\_brightness), global luminance, image contrast, and error rate.

The Character Error Rate (CER) was calculated using the same method used by the ICDAR competition. The CER was determined based on the Levenshtein distance between the recognised text and the ground truth, using the formula shown in equation~\ref{eq}

\begin{equation}
\label{eq}
   \text{ER} = \frac{S + D + I}{N} 
\end{equation}

Initial testing (an example shown in Figure \ref{textdetected}) indicated that the combination of EAST and CRNN achieved a higher average performance, with an CER score of 0.65, while EAST and PyTesseract achieved a CER score of 0.82. Notably, EAST with CRNN performed consistently better across all images, with the exception of those captured at a distance of 1 m with a resolution of 1408×1408 pixels, where a drop in performance was observed.

\begin{figure}[t]
\centering
\includegraphics[width=\textwidth]{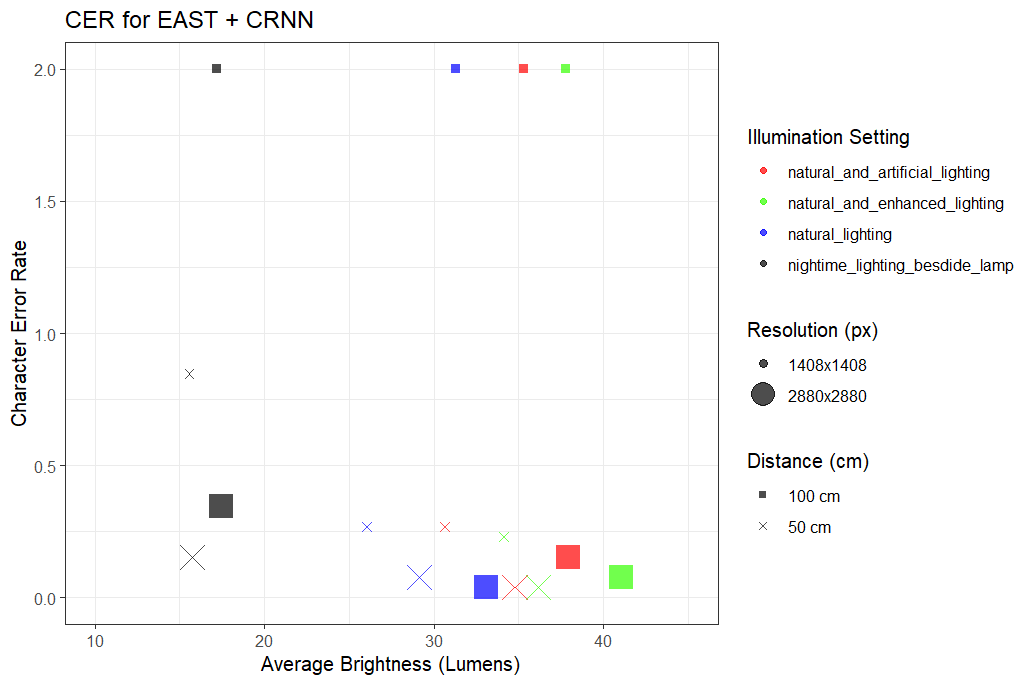}
\caption{CER performance for EAST+CRNN under different environmental conditions.} 
\label{eastcrnn}
\end{figure}

For the EAST and CRNN combination (see Figure \ref{eastcrnn}), resolution appeared to be the most significant factor affecting character recognition rates, showing a strong negative correlation: as resolution decreased, recognition performance deteriorated. Distance was also a substantial influencing factor, with CRNN demonstrating significantly better performance on closer images. In particular, CRNN consistently failed to detect any text on low-resolution (1408 $\times$ 1408) images captured at a 1 m distance, irrespective of lighting conditions, despite bounding boxes being correctly detected. This consistent failure highlights the CRNN high sensitivity to resolution and distance.

\begin{figure}[t]
\centering
\includegraphics[width=\textwidth]{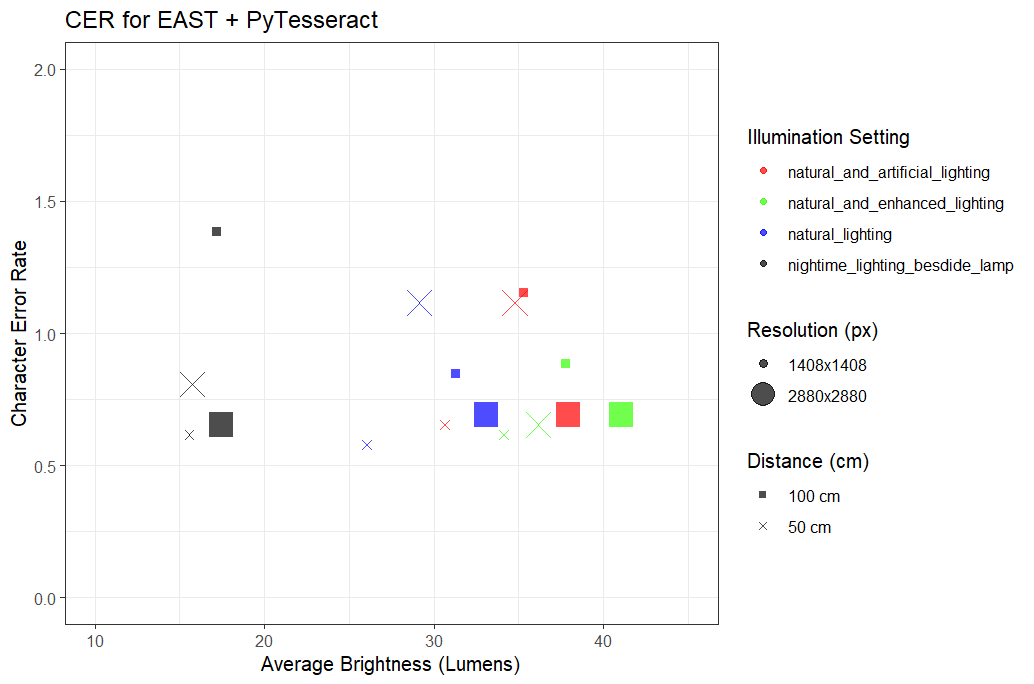}
\caption{CER performance for EAST+PyTesseract under different environmental conditions.} 
\label{eastpytesseract}
\end{figure}

In contrast, for the EAST with PyTesseract combination (see Figure \ref{eastpytesseract}), there was a weaker correlation between resolution, distance and error rate. PyTesseract demonstrated a more consistent performance across varying distances and resolutions. Surprisingly, a slight negative correlation (-0.08) was observed between resolution and error rate. Distance remained a factor, with error rates increasing slightly with greater distance.

Lighting variables impacted both algorithms, but no strong, consistent relationships could be established across all tested conditions (all correlations |r| $\leq$ 0.20 ). For both recognition algorithms, a weak inverse correlation was observed between mean brightness and error rate, indicating that higher brightness could slightly impact recognition accuracy. Interestingly an increase in contrast tended to negatively affect performance across both combinations similarly to the evaluation of EAST.

Overall, PyTesseract appeared less sensitive to changes in environmental variables such as distance, resolution, and lighting. Its performance was more stable across the dataset, although its overall accuracy was lower compared to CRNN. While CRNN achieved higher overall performance, it demonstrated greater vulnerability to variations in resolution and distance.

After a first evaluation, CRNN remains the preferred candidate for this project due to its superior recognition performance. If preprocessing methods can be developed to mitigate its sensitivity to image resolution and distance, CRNN would offer the best overall solution. If such mitigation proves unfeasible, a combined strategy employing both CRNN and PyTesseract based on environmental conditions may be considered for final deployment.

\subsubsection{Improving algorithm results based on environmental findings\\}
\label{sec:improving}

Based on the findings of the first validation, the next step of this project is to improve STDR algorithm performance.

To address the performance issues observed with EAST and CRNN at lower resolutions and greater distances, a preprocessing step involving image upscaling was introduced. All images with a resolution of 1408 $\times$ 1408 were upscaled to 2816 $\times$ 2816 prior to detection and recognition. This adjustment led to a substantial improvement, reducing the overall character error rate from 0.65 to 0.48.

Further refinement was attempted by selectively increasing the brightness of darker images within the dataset. This additional preprocessing step had the opposite effect, increasing the character error rate even further, achieving an average of 0.67 across the evaluated set for the EAST and CRNN pipeline.

Despite EAST and CRNN’s sensitivity to resolution and distance, the application of upscaling lower quality images at a greater distance significantly improved its recognition performance. These results indicate that CRNN, when supported by appropriate preprocessing techniques, can effectively overcome its previously mentioned weaknesses. Therefore, the EAST plus CRNN combination was selected for further evaluation and real-world testing phases.

\subsection{Qualitative validation in the wild with gaze tracking}
\label{sec:exp3}

\begin{figure}[t]
\includegraphics[width=\textwidth]{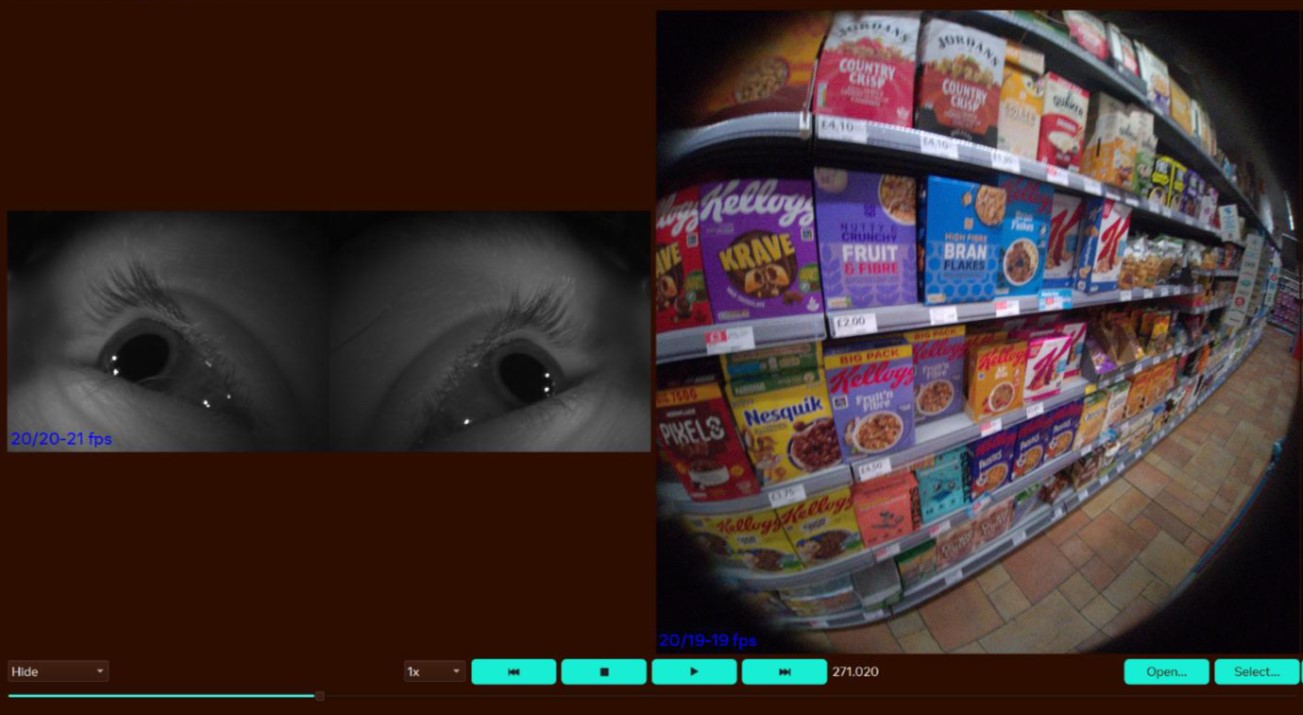}
\caption{A sample of the participant's gaze (left) in parallel with image capture inside a local supermarket (right).} 
\label{sidetoside}
\end{figure}

\begin{figure}
\centering
\begin{subfigure}{.48\textwidth}
  \centering
  \includegraphics[width=\linewidth]{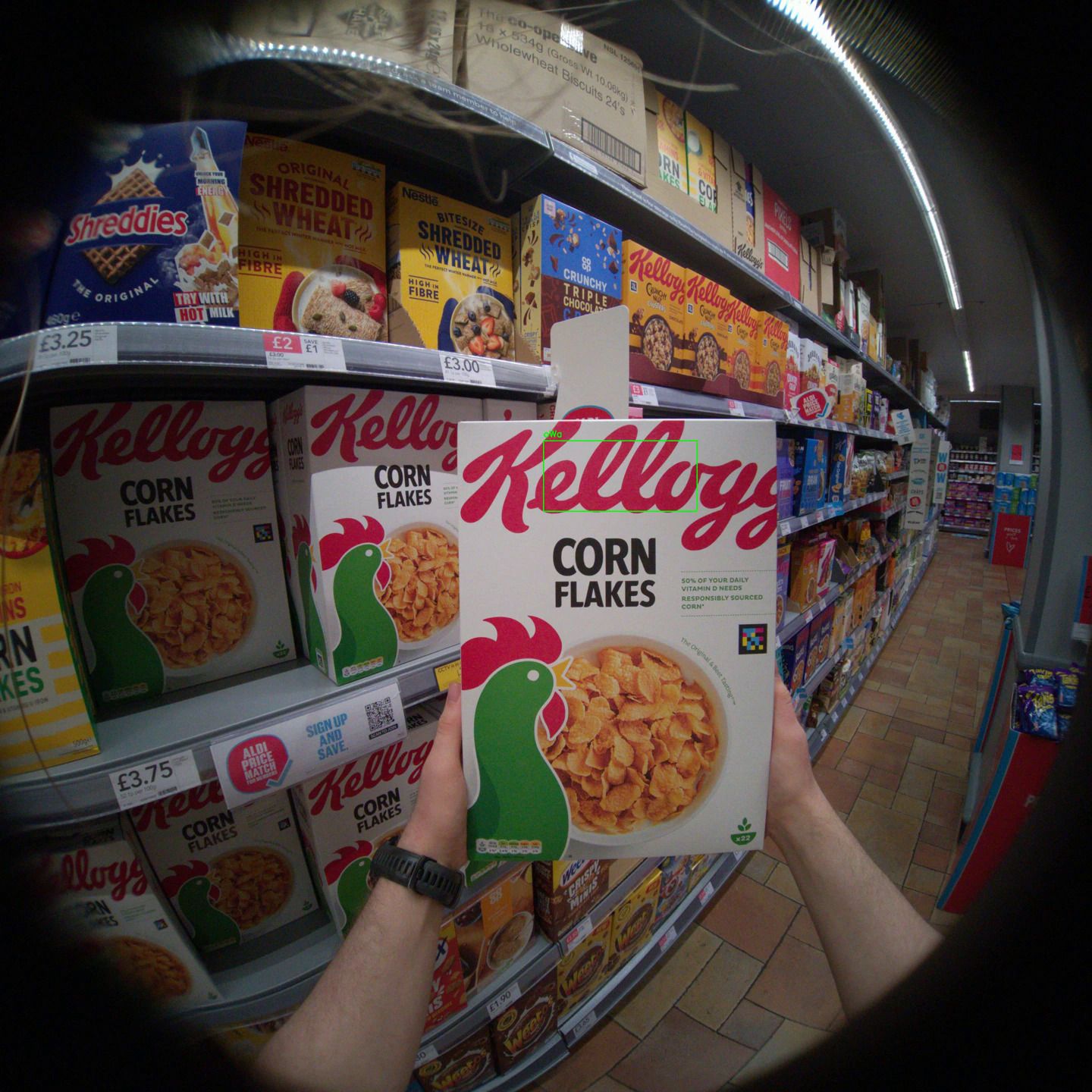}
  \caption{Eye-gaze pointing slightly upwards, detecting the product's logo.}
  \label{fig:sub1}
\end{subfigure}%
\hfill
\vrule width 0.5pt
\hfill
\begin{subfigure}{.48\textwidth}
  \centering
  \includegraphics[width=\linewidth]{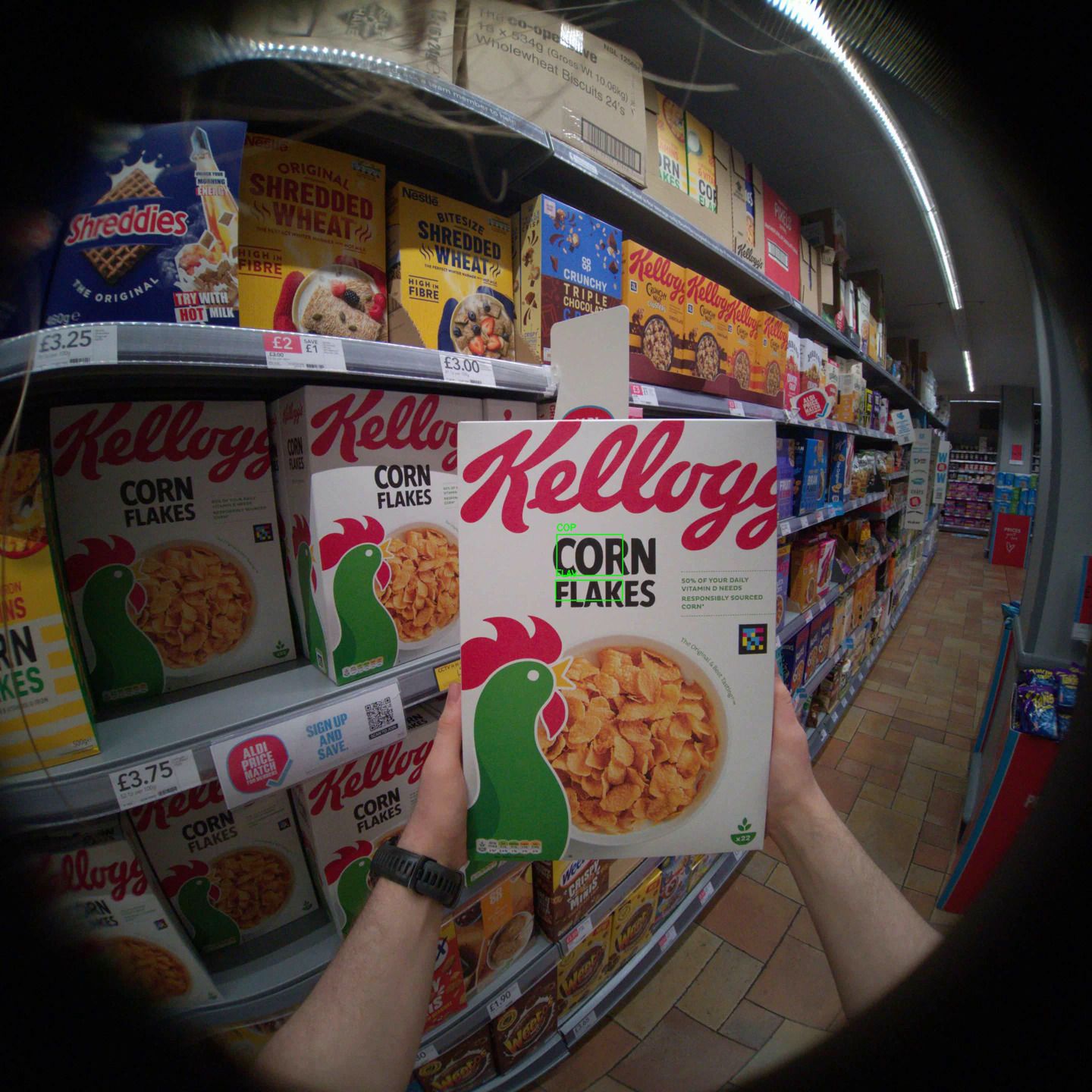}
  \caption{Eye-gaze pointing slightly downwards, detecting the product's name.}
  \label{fig:sub2}
\end{subfigure}
\caption{Different bounding boxes being found depending on the gaze direction.}
\label{cereal_bb}
\end{figure}

For real-world testing, a recording was conducted in a supermarket environment using the Project Aria Glasses (see Figure \ref{sidetoside}). The RGB camera was configured to capture footage at a resolution of 2880×2880 pixels with a frame rate of 20 frames per second. Additionally, the eye-tracking cameras were activated to enable Aria’s eye-tracking services.
Following the recording, the footage was uploaded to Project Aria’s servers to retrieve MPS eye-gaze data. This process took approximately five minutes and involved transferring personal data to remote servers, rendering it unsuitable for real-time processing. However, when used for a demo project and studies not requiring real-time responsiveness, it provided highly accurate and valuable gaze information.
After receiving the eye-gaze data, the EAST-CRNN pipeline was applied selectively. Rather than processing the entire image frame, text detection and recognition were confined to a square area centred around the user’s gaze point, with a diameter approximately one-sixteenth (1/16) of the full image width (see Figure \ref{cereal_bb}).

This approach significantly reduced computation time by focusing only on the region of interest where the user was actively looking, optimising processing efficiency. 
Although the current method is limited by the time required to retrieve eye-gaze data and process frames offline, the results show potential. The system successfully recognised product labels and packaging information in real-world conditions.
It is important to mention however, that this approach only works within the scope of this project. When increasing the text box area to larger scales, the bounding box heuristic failed to accurately group text regions. This is due to the high text density naturally present in such scenarios. This indicates the need to modify this heuristic depending on specific use cases.

%%%%%%%%%%% keeping original just in case we want to go back to previous image
% \begin{figure}
% \centering
% \begin{subfigure}{.5\textwidth}
%   \centering
%   \includegraphics[width=1\linewidth]{cereal_up.jpg}
%   \caption{eye-gaze pointing slightly upwards, detecting the product's logo.}
%   \label{fig:sub1}
% \end{subfigure}%
% \begin{subfigure}{.5\textwidth}
%   \centering
%   \includegraphics[width=1\linewidth]{cereal_down.jpg}
%   \caption{eye-gaze pointing slightly downwards, detecting the product's name.}
%   \label{fig:sub2}
% \end{subfigure}
% \caption{Different bounding boxes being found depending on the gaze direction.}
% \label{cereal_bb}
% \end{figure}

%%%%%%%%%%%%%%%%%%%%%%%%%%%%%%%%%%%%%%%%%%%%%

\section{Conclusions}
\label{concs}
The goal of the evaluation was to determine how various pre-processing techniques could mitigate issues related to environmental conditions, such as lighting, and recording conditions, including resolution and distance from text. It was anticipated that effective pre-processing would enhance text recognition accuracy under these challenging conditions.
However, the results did not meet expectations in relation to lighting. No lighting variable demonstrated a strong or consistent link with the error rate for the EAST-CRNN, or EAST- Pytesseract pipelines. Even when some correlations were observed, applying pre-processing techniques aimed at correcting lighting conditions (such as enhancing brightness) had negative effects on the overall text recognition performance. 
The initial assumption was that this increase in error rate after preprocessing was due to overprocessing already bright images. However, after applying pre-processing only to low-light images, overall performance still worsened.

This outcome highlights an important limitation: correlation does not necessarily imply causation. While certain environmental factors appeared to be linked to increased error rates, modifying them through pre-processing did not consistently improve recognition outcomes.
Several factors could explain this. The impact of lighting variations may be too subtle or complex for simple pre-processing techniques to correct effectively, particularly when utilising pre-trained algorithms. Furthermore, even if positive results were found, the small dataset size means that observed correlations may only apply to specific samples rather than representing generalisable trends.

In contrast, a strong and clear relationship was identified between image resolution, distance from the text, and performance, particularly for the EAST algorithm. These factors had a far more significant impact on recognition accuracy compared to lighting conditions, leading to a reduction in character error rate from 0.65 to 0.48 when using the EAST–CRNN pipeline after upscaling.

Despite the challenges faced during the study, the project and the dataset produced have successfully established a solid base for future research, offering a resource to further investigate optimal methods for robust text recognition from wearable device recordings. It also highlights critical factors, such as resolution and distance, which must be considered when developing or adapting text recognition systems for similar data types.

In hindsight, the lighting variables considered, such as global luminance and mean brightness, were strongly interlinked. This limited the ability to isolate the effect of any single factor. For example, as global luminance increased, mean brightness naturally rose as well, making it difficult to assess their individual impacts on text recognition performance. Consequently, conducting separate studies on each lighting variable was impractical. Future work should instead prioritise testing additional scene-related variables, such as text orientation and background complexity.

Though the project achieved its objectives, there were limitations. These include: the small size of the custom dataset, the computational cost of testing on the ICDAR dataset, and the need to offload data to external servers in order to use proprietary eye gaze retrieval algorithms thus restricting real-time applications. Additionally, bounding box merging was based on heuristic approaches, providing opportunities for refinement.
Several directions for future research are recommended:

\begin{itemize}
    \item Expand the dataset with more varied lighting and complex backgrounds to strengthen statistical analysis.
    \item Prioritise testing additional scene-related variables, such as text orientation, and background complexity.
    \item Investigate real-time integration of gaze-based targeting with lightweight STR models.
\end{itemize}

Overall, the project provides a strong foundation for further development of robust, user-focused STR systems for AR devices.

\end{document}